\documentclass[runningheads]{llncs}
\usepackage{graphicx}

\usepackage{amsmath}
\usepackage{bm}
\usepackage{amsfonts}
\usepackage{amssymb}
\usepackage[nospace]{cite}
\usepackage{url}
\usepackage{xcolor}
\usepackage{fancyhdr}
\usepackage[hidelinks]{hyperref}

\usepackage{adjustbox}
\usepackage{graphicx}
\usepackage{tensor}
\usepackage{tabularx}

\usepackage[utf8]{inputenc}
\usepackage[OT1]{fontenc}
\usepackage[final]{microtype}
\usepackage[nolist,nohyperlinks]{acronym}
\usepackage{multirow}
\usepackage{makecell}
\usepackage{float}

\usepackage{booktabs}

\usepackage[binary-units]{siunitx}

\usepackage{lipsum}

\DeclareMathAlphabet{\pazocal}{OMS}{zplm}{m}{n}

\newcommand{\Ds}{\pazocal{D}}

\newcommand{\Es}{\pazocal{E}}

\newcommand{\Ls}{\pazocal{L}}

\newcommand{\Ms}{\pazocal{M}}

\newcommand{\Dcoll}{\mathbb{D}_{coll}}
\newcommand{\D}{\mathbb{D}}

\newcommand{\z}{\mathbf{z}}
\newcommand{\x}{\mathbf{x}}
\newcommand{\xcoll}{\mathbf{x}_{coll}}
\acrodef{collnet}[DCE]{Depth image-based Collision Encoder}

\begin{document}
\title{Task-driven Compression for Collision Encoding based on Depth Images}

\author{Mihir Kulkarni and Kostas Alexis}
\authorrunning{Mihir Kulkarni and Kostas Alexis}
\institute{Norwegian University of Science and Technology (NTNU), O. S. Bragstads Plass 2D, 7034, Trondheim, Norway \\
\email{mihir.kulkarni@ntnu.no}}

% \institute{Norwegian University of Science and Technology, Norway}
%
\maketitle              % typeset the header of the contribution
\begin{abstract}
This paper contributes a novel learning-based method for aggressive task-driven compression of depth images and their encoding as images tailored to collision prediction for robotic systems. A novel 3D image processing methodology is proposed that accounts for the robot's size in order to appropriately ``inflate'' the obstacles represented in the depth image and thus obtain the distance that can be traversed by the robot in a collision-free manner along any given ray within the camera frustum. Such depth-and-collision image pairs are used to train a neural network that follows the architecture of Variational Autoencoders to compress-and-transform the information in the original depth image to derive a latent representation that encodes the collision information for the given depth image. We compare our proposed task-driven encoding method with classical task-agnostic methods and demonstrate superior performance for the task of collision image prediction from extremely low-dimensional latent spaces. A set of comparative studies show that the proposed approach is capable of encoding depth image-and-collision image tuples from complex scenes with thin obstacles at long distances better than the classical methods at compression ratios as high as $4050:1$.

\end{abstract}

\section{Introduction}

Methods for autonomous collision-free navigation of aerial robots have traditionally relied on motion planning techniques that exploit a dense map representation of the environment~\cite{tordesillas2019faster,GBPLANNER_JFR_2020,voxblox,rocha2023overview}. Departing from such methods, the community has recently investigated the potential of deep learning to develop navigation methods that act directly on exteroceptive data such as depth images instead of reconstructed maps in order to plan the aerial vehicle's motions with minimal latency~\cite{Loquercio2021Science,loquercio2023agile,ORACLE_ICRA2022,sevae_oracle_2023}. However, such methods face the challenge that exteroceptive data and especially depth images coming from stereo vision or other sensors are typically of very high dimensionality and the involved neural networks include layers that partially act as lossy information compression stages. This is reflected in the architectures of otherwise successful methods such as the works in~\cite{Loquercio2021Science,ORACLE_ICRA2022,sevae_oracle_2023} that exploit depth images to evaluate which among a set of candidate robot trajectories would collide or not. In~\cite{Loquercio2021Science} the input depth image involves more than $300,000$ pixels ($640\times 480$ resolution) but through stages of a pre-trained MobileNetV3 architecture it gets processed to $M$ feature vectors of size $32$ each, where $M$ is the number of candidate trajectories for which this method derives collision scores. Eventually by combining the $640\times 480$ pixels depth image with robot pose information, the method attempts to predict which among $M$ trajectories are safe, thus representing a process of information downsampling and targeted inference. In other words, despite the dimensionality reduction taking place through the neural network it is attempted that the method still ensures collision avoidance. However, it is known that such techniques do not provide $100\%$ success ratio especially in complex and cluttered scenes. 

\begin{figure}
    \vspace{-4ex}
    \centering
    \includegraphics[width=\columnwidth]{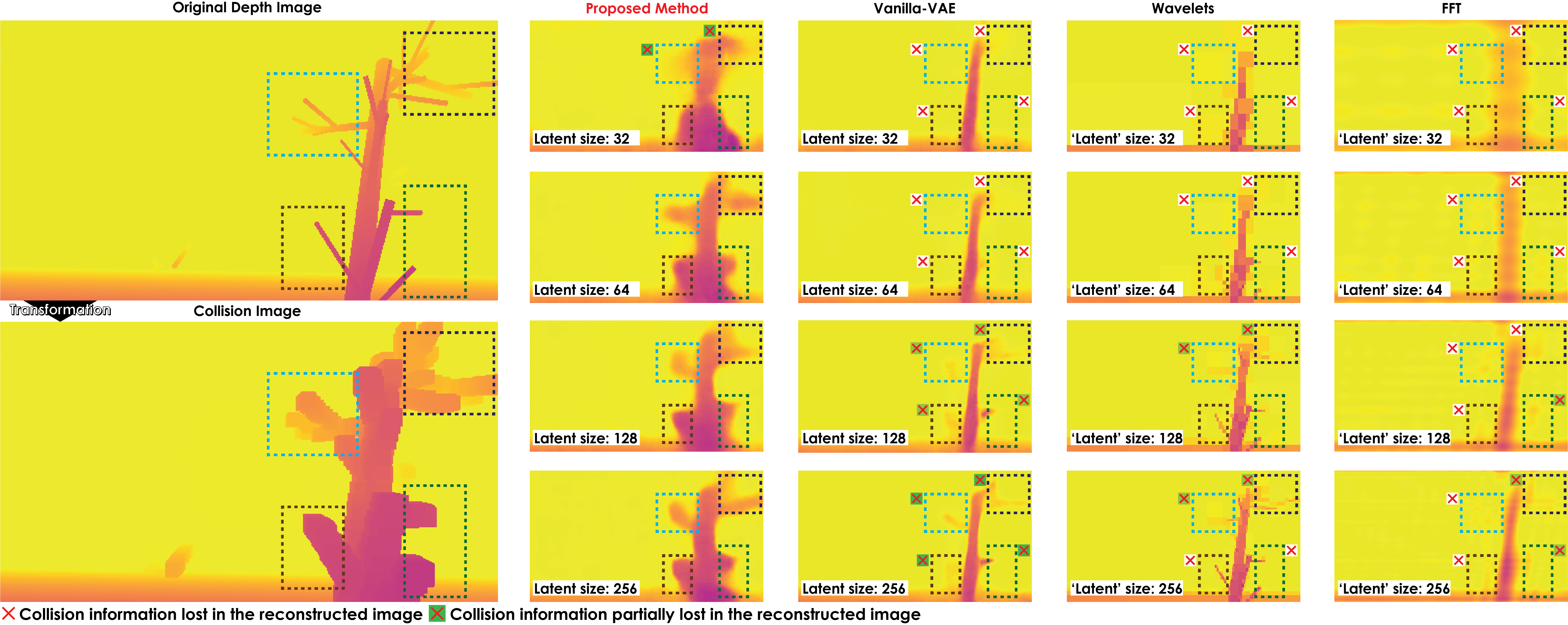}
    \vspace{-4ex}
    \caption{Aggressive compression/encoding of depth images on aggresively low-dimensional latent spaces using conventional techniques is likely to lead to major loss of collision information. On the contrary, a task-driven compression paradigm is proposed that allows to retain most of the collision information even in exceptionally low latent spaces. This work serves as a modular step that delivers compressed latent spaces that retain collision information and can thus be utilized for further processing by methods that predict the possible collision of candidate trajectories of robots in complex scenes. }
    \label{fig:intro_fig}
    \vspace{-4ex}
\end{figure}
% We propose a method that to encode and represent collision information from depth images in highly compressed latent representations. We compare our approach against traditional image compression methods and a task-agnostic VAE to show and demonstrate superior capabilities for encoding and reconstructing features in complex and cluttered environments.

Responding to the above, this work contributes the concept of task-driven compression and encoding of depth images as visualized in Figure~\ref{fig:intro_fig}. Departing from the concept that methods aiming to predict the safety of candidate robot trajectories based on depth images should train collision prediction either a) directly in an end-to-end fashion through depth data~\cite{Loquercio2021Science,ORACLE_ICRA2022} or through b) an explicit intermediate compression stage of the depth image itself~\cite{niu2023embarrassingly}, we propose the approach of using the depth image to encode a latent space presenting major dimensionality reduction that reflects not the depth image itself but instead a ``collision image''. The latter is a remapping of the depth image that has accounted about the robot's size and thus presents reduced overall complexity and greatly reduced presence of narrow/thin features that are hard-to-retain in an aggressive compression step. To achieve this goal, the method employs a probabilistic encoder-decoder architecture that is trained in a supervised manner such that given a depth image as input, it learns to encode and reconstruct the collision image. To train this collision-predicting network --dubbed \ac{collnet}-- the depth image is first processed such that the collision image is calculated given information for the robot's size. Focusing on aggressive dimensionality reduction, it is demonstrated that the scheme allows to get accurate reconstructions through a latent space that is more than $3$ orders of magnitude smaller than the input image. The benefits of the approach are demonstrated through comparisons both with a conventional Variational Autoencoder (VAE) trained to encode the depth image and assessed regarding the extent to which the reconstructed image can serve as basis to derive a correct collision image, as well as traditional compression methods using the Fast Fourier Transform (FFT) and wavelets.

%Motivated by the above, this work explicitly considers the problem of a) delivering the necessary dimensionality reduction of depth images acquired aerial robots to be used for predicting if a certain candidate path for the robot shall be in collision with the environment, while b) ensuring 

In the remaining paper Section~\ref{sec:related_work} presents related work and Section~\ref{sec:proposed_method} details the proposed method involving generation of training data, image augmentation and the training of the neural network. Section~\ref{sec:results} compares our proposed method against traditional image compression methods and evaluates the performance of task-driven and task-agnostic compression methods at similar degrees of compression. Finally, conclusions are drawn in Section~\ref{sec:conclusions}.

\section{Related Work} \label{sec:related_work}

This work draws its motivation from the set of deep learning methods that rely on directly processing sensor data (such as depth images) in order to predict if a candidate trajectory of a flying robot shall be in collision or not~\cite{loquercio2023agile,niu2023embarrassingly,Loquercio2021Science,ORACLE_ICRA2022,sevae_oracle_2023} and accordingly enable safe autonomous navigation. A subset of such methods instead of relying on direct end-to-end learning from exteroceptive data and robot pose information to predict if a certain candidate action/trajectory shall allow collision-free flight, they employ modularization and accordingly an explicit step of compression that pre-processes high-dimensional input image data arriving to a low-dimensional latent space~\cite{sevae_oracle_2023,niu2023embarrassingly}. 

Technically, the contribution relates to the body of work on image compression. In this large body of work, multiple methods are available including classical schemes that rely on FFT or wavelets~\cite{dhawan2011review,lewis1992image}. Within the breadth of relevant techniques, of special interest is the utilization deep learning approaches~\cite{mishra2022deep,cheng2019deep} and especially variational autoencoders~\cite{higgins2017betavae,doersch2016tutorial,pu2016variational} as means to achieve good reconstruction quality for high compression ratios~\cite{zhou2018variational,wen2019variational}. Nevertheless, the majority of such methods follow the main paradigm of compression which implies that a uniform metric (e.g., mean squared loss) of over pixel-level reconstruction against the original image is employed. Even for works that exploit additional cues such as semantics~\cite{wang2019end}, conventional compression remains the prime goal. Departing from this paradigm this work reflects the fact that in the line of works of collision prediction~\cite{loquercio2023agile,niu2023embarrassingly,Loquercio2021Science,ORACLE_ICRA2022,sevae_oracle_2023} it is the information over candidate collisions that matters and not the depth pixels themselves. In other words, it is the question if the robot - with the specific volume that it occupies - can fly along a path within the volume observed and captured by the depth image. This calls for a new concept that hereby is called purposeful task-driven depth image compression/encoding for collision prediction utilizing minimal latent spaces. It is highlighted that the goal to arrive at a latent space that is multiple orders of magnitude smaller than the high dimensional depth images --offered by sensing solutions such as modern stereo vision-- is driven from the need of robust performance and generalization in diverse natural environments. As established by seminal works such as ResNet~\cite{he2016deep}, deeper models with more parameters require much more data to train. A low-dimensional compression latent space enables methods that shall then use it for collision prediction~\cite{sevae_oracle_2023} to utilize smaller and simpler networks for the task, while they further combine with robot data which are also low-dimensional (e.g., pose states of a quadrotor aerial vehicle over the SE(3) special Euclidean group~\cite{lee2013geometric}).

\section{Proposed Method} \label{sec:proposed_method}

The proposed approach on task-driven compression and particularly depth image-based collision encoding is outlined below. First, the process to generate relevant training data is discussed, followed by the method to derive the collision image associated with each depth image. Subsequently, the depth images-based collision encoder motivated by the architecture of variational autoencoders is presented. 

\subsection{Dataset Generation} \label{sec:data_acquisition}
Deep learning techniques for data compression require large amounts of data for training. Moreover, the generalizability of the learned models depends on the quality of the training data and the variety of samples provided for learning. Available depth image datasets primarily focus on specific tasks to be performed using the depth images such as depth completion~\cite{silberman2012nyu_depth} or autonomous driving~\cite{packnet}. These datasets contain images from scenes that include urban structured indoor settings and open streets respectively with large-sized obstacles that are sparsely distributed in the environment. Consequently, such datasets - that are otherwise common within both research and industry - do not contain images from highly cluttered complex environments that present challenges to aerial robot navigation. For the latter, it is important to note that environments with a) high clutter leading to uncertainty as to the safest flying direction, and b) obstacles with narrow cross section (``thin'' obstacles) are particularly hard to fly through. In order to train our neural network models for such cluttered environments containing narrow/thin obstacles, while ensuring generalizability, we rely on two popular robot simulators - namely Gazebo Classic~\cite{gazbeoclassic} and Isaac Gym~\cite{isaac_gym} to generate diverse simulated depth image data. These simulators provide the necessary interfaces that allow us to rearrange different objects randomly in a simulated environment. Images from Gazebo Classic are collected using the onboard depth camera of a simulated aerial robot in an obstacle-rich environment using the RotorS Simulator~\cite{furrer2016rotors}. Subsequently, we utilize the Isaac Gym-based Aerial Gym Simulator~\cite{kulkarni2023aerialgym} in order to simulate environments with randomly placed obstacles and collect depth images in a parallelized manner from multiple randomly generated environments simultaneously. $85,000$ depth images are collected in environments consisting of a variety of objects ranging from multi-branched tree-like objects with thin cross-sections to large obstacles with cavities in them. Depth images are collected and aggregated to be processed for computing a robot-specific collision image.

\subsection{Collision Image Generation} \label{sec:collision_image_generation}
While the collected depth images provide information about the projected distance to a surface along the central axis of the camera, it is difficult to infer the collision-free regions in the robot's field-of-view. Traditional approaches to compute collision-free regions involve representing the depth image in an intermediate volumetric map-based representation~\cite{hornung13auro, voxblox, museth2013vdb} that can be queried to derive collision-free regions. These representations are limited by their discretization capabilities and often require a large amount of memory to maintain a persistent map~\cite{voxblox}. Generation of such representations is also a computationally expensive step~\cite{hornung13auro}. Finally, such reconstructions rely on aggregating multiple depth image readings and thus necessitate consistent pose estimation. At the same time, methods that use depth images to directly predict if a candidate path is collision-free or not~\cite{ORACLE_ICRA2022,Loquercio2021Science} implicitly have to learn that the depth image itself is not a map of collision-free space but instead this information can be acquired by further correlating the range to a point and the size of the robot. Contrary to the current techniques on that front that typically either a) resort on end-to-end learning of collisions via depth, state and action tuples~\cite{ORACLE_ICRA2022,Loquercio2021Science} or b) compress the depth image and use this lossy latent space to then learn collision prediction~\cite{sevae_oracle_2023,niu2023embarrassingly}, we here propose the re-mapped representation of depth images in a new form that directly provides the collision-free distance that can be traversed by a robot along any direction. A collision image is defined as an image representing the collision-free distance (projected along the central axis of the camera) traversable by a robot of known dimensions along the rays corresponding to each pixel in an image. This revised image representation that encodes all necessary collision information can then be utilized directly for robot navigation tasks. 

\begin{figure}
    \centering
    \includegraphics[width=0.91\columnwidth]{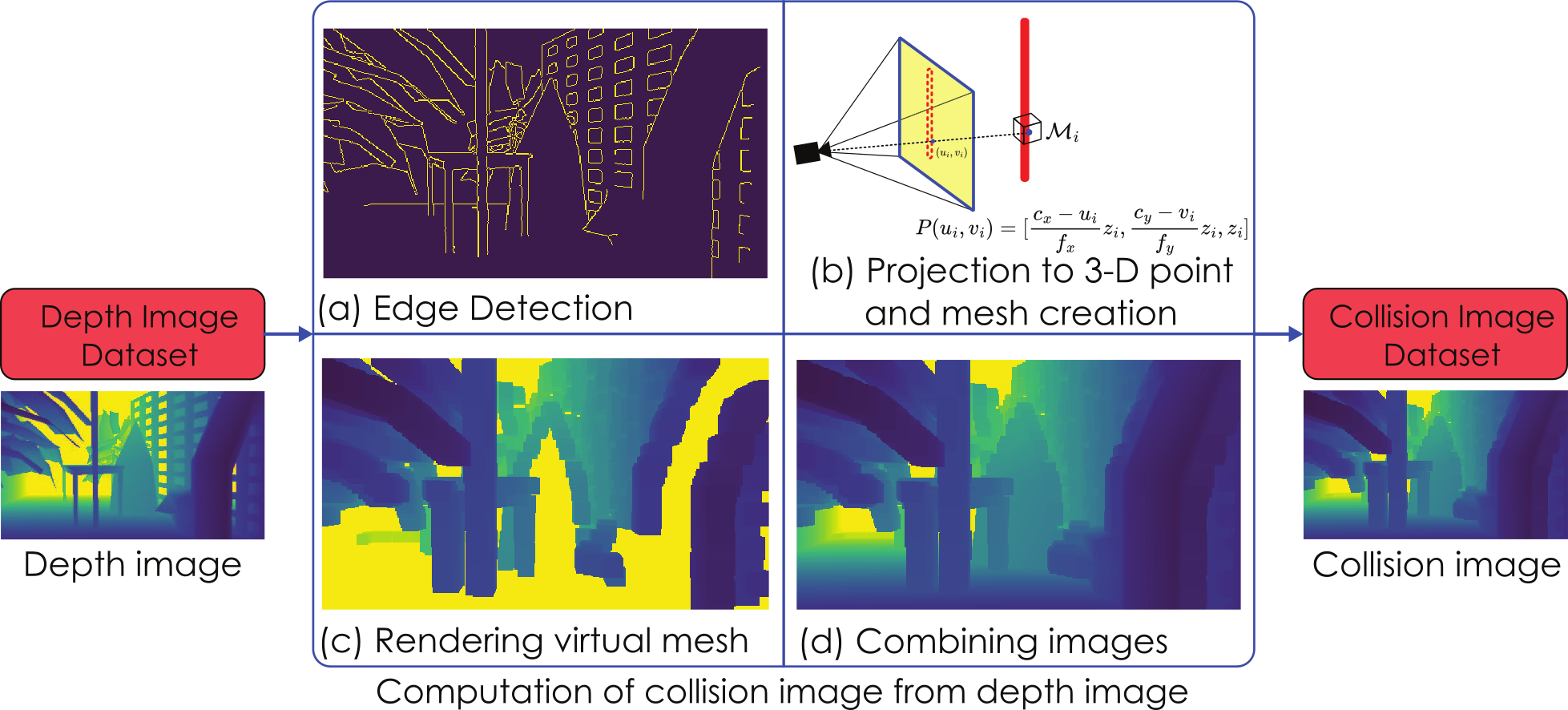} %DatasetProcessing
    \vspace{-2ex}
    \caption{The acquired dataset is processed for task-driven compression. Edge-detection is performed on the depth images and each edge pixel is projected to it's $3$D coordinates to form a pointcloud and a virtual $3$D mesh is rendered. The depth image of the virtual mesh is obtained and combined with an offset range image to form a collision image.}
    \label{fig:dataset_processing}
    \vspace{-3ex}
\end{figure}

To derive collision images from depth images, we propose a computationally efficient method illustrated in Figure~\ref{fig:dataset_processing}. Motivated by the observation that the most significant change between the depth image and the collision image occurs at the edges of obstacles in the field-of-view of the camera, a rendering-based approach is utilized to appropriately inflate the objects in the camera's field-of-view about their edges. We cannot perform this inflation accurately using traditional $2\textrm{D}$ computer vision techniques since the modified area around each edge pixel is both dependent on the size of the robot and the distance to the point in $3$D space making the computation intractable. We rely on parallelized rendering frameworks to visualize virtual robot-sized meshes around the regions corresponding to the edges of the obstacles in order to inflate them by the size of the robot. Projecting them back onto the camera plane captures the appropriately inflated regions of the environments that represent the regions of collision for the robot. Edge detection is performed on the original depth image $\Ds$ using OpenCV~\cite{opencv_library} to obtain the set of pixels corresponding to the edges $\Es$ as shown in Figure~\ref{fig:dataset_processing}(a). A fraction of the edge pixels are randomly selected to render meshes. For each selected edge pixel $i$ with coordinates $(u_i,v_i) \in \Es$, the position of the corresponding point $P_i \in \mathbb{R}^3$ is calculated as:

\begin{align}
    P_i &= (x_i, y_i, z_i), \\
    \intertext{where}
    x_i &= \frac{c_x - u_i}{f_x} z_i, \\
    y_i &= \frac{c_y - v_i}{f_y} z_i, \\
    z_i &= \Ds(u_i, v_i).
\end{align}

A pinhole model of the camera is considered, with $f_x~\textrm{and}~f_y$ as the focal lengths and $c_x~\textrm{and}~c_y$ as the optical centers. The shape of the robot is considered to be cubical with edge length $2r$. For each projected point $P_i$, a robot-sized mesh $\Ms_i$ is centered at the coordinates $(x_i, y_i, z_i)$ as shown in Figure~\ref{fig:dataset_processing}(b). Meshes created around each point are merged into a single aggregated mesh $\Ms$. We use NVIDIA Warp~\cite{warp2022}, a high-performance graphics and simulation package that enables rendering simulated depth cameras in a virtual environment consisting of this aggregated mesh. A parallelized ray-casting operation is performed to project rays into this virtual mesh environment and obtain a depth image $\Ds_\Ms$ only containing these virtual meshes (Figure~\ref{fig:dataset_processing}(c)). This depth image only contains the information regarding the distances to the virtual meshes corresponding to the edge pixels in the original depth image $\Ds$. Since rendering of virtual meshes is a computationally expensive step, it is reserved only for the edge pixels in the image. For pixels lying in the interior regions of the object in the depth image, an offset depth image $\Ds_{\textrm{offset}}$ is created with all range values brought closer by the size of the robot $r$ using the following operation:

\begin{align}
    \Ds_{\textrm{offset}} &= \mathcal{R}^{-1}(\mathcal{R}(\Ds) - r),
\end{align}
where the transformation $\mathcal{R}$ converts the depth image to a range image, i.e., the value in each pixel of the image represents the Euclidean distance to the corresponding point on the object. The inverse function $\mathcal{R}^{-1}$ converts the range image back to a depth image. Finally, an approximate collision image $\Ds_{\textrm{coll}}$ is obtained by taking pixel-wise minimum values of the offset depth image $\Ds_{\textrm{offset}}$ and rendered image with inflated meshes $\Ds_{\Ms}$ as shown in Figure~\ref{fig:dataset_processing}(d). This operation is given by: 
\begin{align}
    \Ds_{\textrm{coll}} &= min(\Ds_{\Ms}, \Ds_{\textrm{offset}}).
\end{align}

We use this to generate a collision image dataset given the depth image dataset with each image in the original dataset being processed in the above manner to produce a collision image. Both the original image and the collision image are aggregated into a common dataset to be used for training the probabilistic encoder-decoder network to derive and encode the collision information from the original depth images.

\subsection{Depth Image Compression and Collision Encoding} \label{sec:vae}
The interpretation and representation of depth information to derive collision images requires spatial understanding of the environment. We utilize artificial neural networks to perform this task by learning a compressed representation that compresses and encodes the depth image to its associated collision image. The overall architecture is motivated by the success of VAEs but with the important distinction that the involved learning includes training of the depth-to-collision image map transformation. We consider a dataset containing depth images $\x \in \D$, and its derived secondary dataset containing collision images $\xcoll \in \Dcoll$. A surjective function $\mathcal{P}: \D \mapsto \Dcoll$ maps each element from the depth image dataset to an image in the collision image dataset. This function is imitated in the collision image generation step (Section~\ref{sec:collision_image_generation}). Each $\xcoll \in \Dcoll$ can be assumed to be generated by a process using a latent random variable $\z$.

\begin{figure}
    \vspace{-2ex}
    \centering
    \includegraphics[width=0.92\columnwidth]{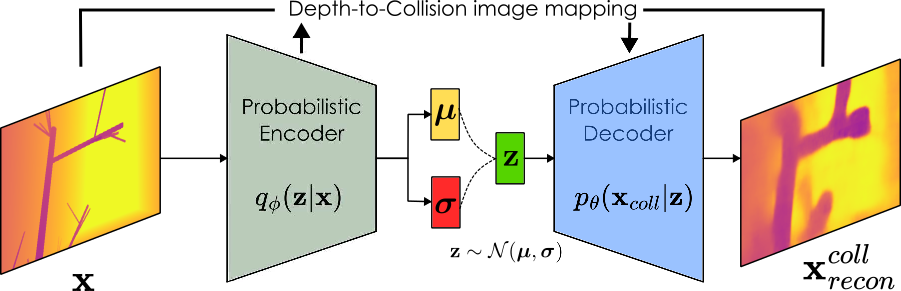}
    \caption{The proposed neural network with an encoder-decoder architecture inspired by variational autoencoders and tailored to compress and re-map a depth image $\x$ to a latent representation $\z$ that can be used to produce the reconstructed image $\x^{coll}_{recon}$ that approximates the associated collision image $\xcoll$.}
    \label{fig:nn_architecture}
    \vspace{-4ex}
\end{figure}

 We employ probabilistic encoders and decoders to perform dimensionality reduction of the input depth data and learn a highly compressed latent representation for predicting collision images. A probabilistic decoder $p_{\theta}(\xcoll| \z)$, given $\z$ produces a distribution over the possible values of $\xcoll$, while a probabilistic encoder $q_{\phi}(\z| \x)$ learns to encode the input image $\x$ to a latent distribution with mean $\boldsymbol{\mu}$ and standard deviation $\boldsymbol{\sigma}$. This distribution is sampled to obtain $\z$ such that $\z \sim \mathcal{N}(\boldsymbol{\mu}, \boldsymbol{\sigma}\cdot\mathbf{I})$. The encoder and decoder networks are jointly trained to produce a highly compressed but well performing latent representation $\z$ given a depth image $\x$ and its $\xcoll$. The decoder can be used to derive a collision image that approximates $\xcoll$ and accurately predicts the distances for collision-free traversal using the given depth image. Figure~\ref{fig:nn_architecture} shows the structure of the \ac{collnet} for task-driven compression. To train the \ac{collnet} the loss function is defined as:

\begin{align}
    \Ls &= \Ls_{recon} + \beta_{norm}\Ls_{KL},\\
    \intertext{where}
    \Ls_{recon}(\xcoll, \x^{coll}_{recon}) &= \textrm{MSE}(\xcoll, \x^{coll}_{recon}),\\
    \Ls_{KL}(\bm{\mu}, \bm{\sigma}) &= \frac{1}{2}\sum_{j=1}^J \left (1 + \log(\bm{\sigma}_j^2) - \bm{\mu}_j^2 - \bm{\sigma}_j^2 \right).
\end{align}

Here, $\Ls$ denotes the overall loss term while $\Ls_{recon}$ and $\Ls_{KL}$ (scaled by a constant $\beta_{norm}$~\cite{higgins2017betavae}) denote the reconstruction loss and the KL-divergence loss terms respectively in a manner motivated by autoencoder literature~\cite{doersch2016tutorial}. The Mean-Square Error (MSE) loss function is modified to ignore the errors of the pixels from the depth image that are invalid, i.e., the pixels that do not contain accurate depth information owing to the obstacles being too close to the camera in simulated images or also in case of the incorrect depth from stereo shadows for real-world depth images. The encoder is a residual neural network consisting of convolutional layers at each block and uses the ELU activation function. The final layers of the encoder network are fully connected layers that produce the mean and variance describing the latent distribution. The decoder consists of two fully connected layers followed by non-residual de-convolutional layers with ReLU activation functions. The last convolutional layer has a sigmoid activation to have bounded values for the collision image. The network is trained on a dataset consisting of $70,000$ depth and collision image pairs and tested on a dataset containing $15,000$ image pairs. Each image has a dimension of $270\times480$ pixels and contains the distance to the given obstacle projected along the central axis of the camera. As discussed in the next section, well performing latent spaces as low as $32$ variables are achieved which represents more than $3$ orders of magnitude compression, while simultaneously delivering and exploiting the described depth-to-collision image transformation. 

\section{Evaluation and Results} \label{sec:results}

The main premise of the work is that the implicitly learned transformation of depth-to-collision image mapping, not only allows to learn directly the information pertinent to collision prediction, but also allows major compression while retaining the necessary information. To demonstrate this fact, we conduct a comprehensive set of evaluation studies comparing the performance of our proposed approach against traditional task-agnostic compression methods such as using the wavelet transform and the Fast Fourier Transform (FFT). We also compare our task-driven compression method against a conventionally trained task-agnostic VAE (vanilla-VAE) that shares the same neural network architecture as the \ac{collnet}. We first show that neural network-based compression outperforms traditional compression methods such as FFT and wavelet transform-based compression for very high compression ratios for depth images. Furthermore, the reconstructed collision image obtained from the task-driven \ac{collnet} accurately represents the calculated collision image as compared to the derived collision information from the image reconstructed from the vanilla-VAE. The performance of the proposed approach is evaluated for a set of different latent dimensions representing varying levels of extreme compression. Latent spaces of $32,~64,~128~\textrm{and}~256$ latent dimensions corresponding to compression factors of $4050,~2025,~1012.5~\textrm{and}~506.25$ respectively are considered. The proposed learning-based compression and image domain transformation method not only outperform the currently established approaches while achieving large compression ratios but also are capable of encoding spatial information from the depth image to represent collision information. This is made evident from the results where the depth image is accurately (and range- and robot size-dependent) ``inflated'' to obtain a collision image that occludes the obstacles in the background.

\subsection{Comparison of vanilla-VAE with traditional compression methods}

We compare a vanilla-VAE based compression with FFT and wavelet transform-based compression. The task-agnostic vanilla-VAE is trained using $70,000$ images to encode a depth image $\x$ into a latent distribution and also to reconstruct the input depth image $\textbf{x}^{vanilla}_{recon}$. This is done to first ensure a fair comparison between task-agnostic methods. A separate network is trained on the dataset for each latent space size.

\begin{figure}[h]
    \vspace{-2ex}
    \centering
    \centering
    \includegraphics[width=0.95\columnwidth]{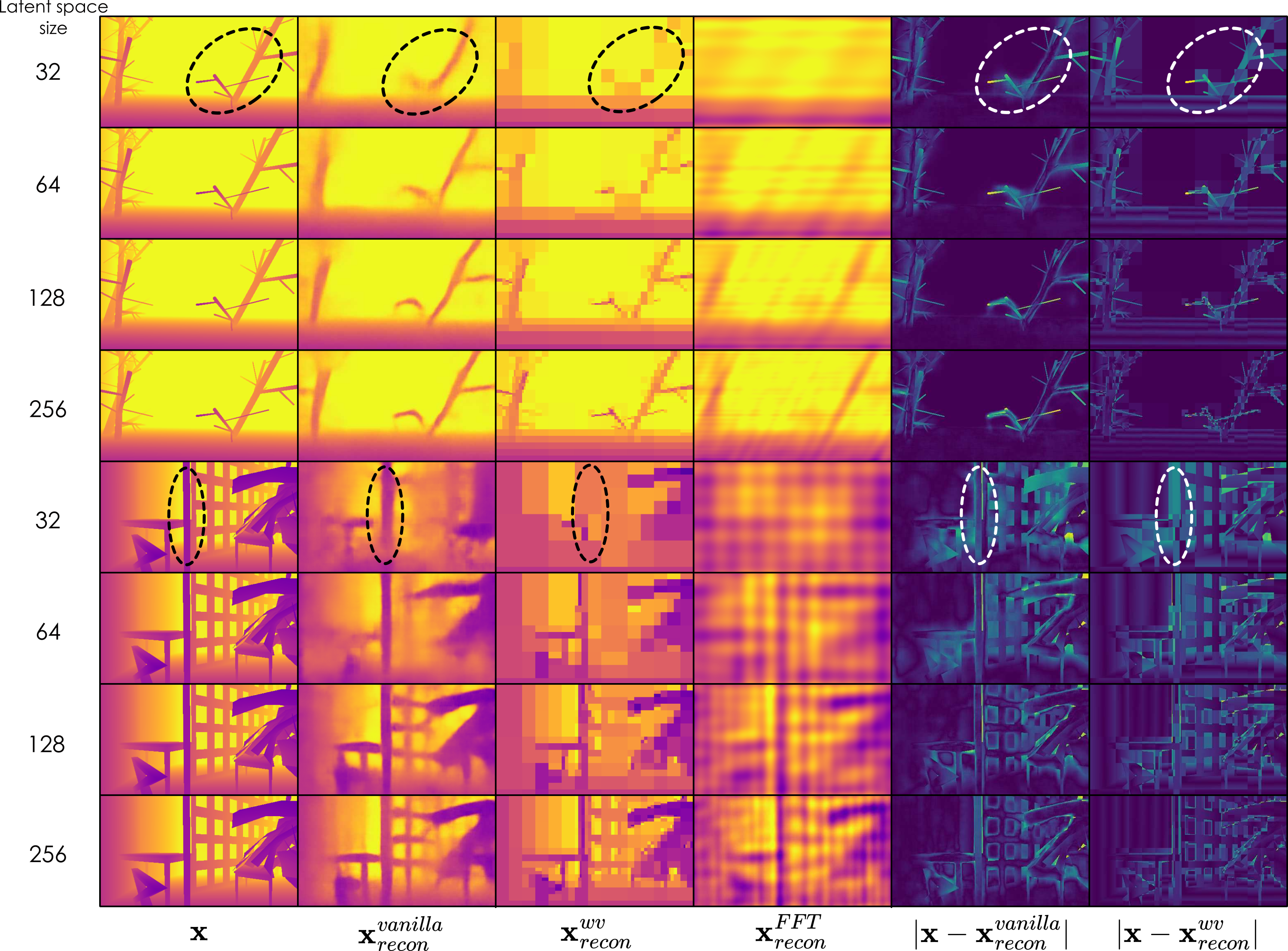}
    \vspace{-2ex}
    \caption{Comparison between the reconstruction performance on depth images using traditional methods and the vanilla-VAE for different levels of compression. The images compressed and reconstructed using vanilla-VAE ($\x^{vanilla}_{recon}$), wavelet transform ($\x^{wv}_{recon}$) and FFT ($\x^{FFT}_{recon}$) are shown. The errors in the reconstruction are also highlighted.}
    \label{fig:vanilla_vae_vs_traditional}
    \vspace{-6ex}
\end{figure}

We obtain the image representation in the wavelet domain by decomposing the image with the Daubechies wavelet `db$1$'. To obtain the compressed representation in this domain corresponding to a latent space size of $n$, the largest $n$ magnitudes in the wavelet domain are retained, while all other values are set to $0$. The resultant wavelet domain representation is reconstructed using the inverse wavelet transform to obtain $\textbf{x}^{wv}_{recon}$. Similarly, to compress the image using FFT, the complex numbers in the frequency domain that correspond to the $n/2$ largest magnitudes are retained (with both their real and complex coefficients), while all others are set to $0$. A reconstruction $\textbf{x}^{FFT}_{recon}$ is obtained from this compressed representation by performing an inverse FFT. It must be noted that both these representations are computationally represented as ordered lists that contain the position-dependent coefficients for the decoder to reconstruct the image. While we retain only the top $n$ coefficients, we do not remove their position information to allow the reconstruction software to work seamlessly. As a result, information retained using this scheme is \textit{more} than just the $n$ dimensional variable that we use in the case of the neural networks.
%It must be noted that that the representations obtained with this compression scheme implicitly contain \textit{more} information corresponding to the position of each of the discrete magnitudes of signal that are retained both for FFT and wavelet transform.
Figure~\ref{fig:vanilla_vae_vs_traditional} compares the reconstructed images from the compressed representation for different latent space sizes using different compression methods. The vanilla-VAE preserves the features in the depth image for complex scenes for small latent sizes, while the wavelet transform-based compression performs well for larger latent space sizes. The difference between the reconstructed image using the vanilla-VAE and the wavelet transforms and the input image is shown to highlight the regions with a higher reconstruction error. A visual inspection of the reconstructed images from wavelets and frequency domain representations show that these methodologies are unable to encode the information in complex depth images for smaller latent space dimensions. The difference is especially highlighted in images that contain complex and cluttered settings, where the FFT reconstructions generate artificial patterns, while the wavelet reconstructions discretize regions of the image non-uniformly, losing out on the sharper details of the image.

The results are tabulated in Table~\ref{tab:comparison_traditional_methods} demonstrating that for high compression ratios corresponding to latent spaces of $32,~64~\textrm{and}~128$ dimensions, the vanilla-VAE based depth image compression method produces images with a lower MSE value with the input image. Interestingly, the wavelet transform-based method produces a lower MSE in the case where the information corresponding to the top $256$ coefficients is retained. As shown in Figure~\ref{fig:vanilla_vae_vs_traditional}, the wavelet reconstruction corresponding to this size produces sharper edges in the reconstructed image owing to the capability to encode more information regarding the smaller discretized regions in the image.

\begin{table}[h]
    \vspace{-4ex}
    \centering
    \label{tab:comparison_traditional_methods}
    \caption{Comparison of MSE for reconstructed images with vanilla-VAE, FFT and wavelet transform for different compressed latent dimensions.}
    \vspace{-2ex}
    \begin{tabularx}{\columnwidth}{|l|X|X|X|X|}
      \hline
      \multicolumn{5}{|c|}{\textbf{MSE against input image $\x$}} \\
      \hline
      \textbf{Latent dims:} & \textbf{32} & \textbf{64} & \textbf{128} & \textbf{256}\\
      \hline
      $\x^{vanilla}_{recon}$ & $\textbf{1249.58}$ & $\textbf{827.00}$ & $\textbf{543.38}$ & $477.88$ \\
      \hline
      $\x^{wv}_{recon}$ & $1481.36$ & $952.58$ & $612.31$ & $\textbf{382.43}$ \\
      \hline
      $\x^{FFT}_{recon}$ & $2223.87$ & $1634.52$ & $1181.93$ & $840.38$ \\
      \hline
    \end{tabularx}
    \vspace{-6ex}
\end{table}

\subsection{Task-driven compression for collision representation}
While the task-agnostic vanilla-VAE demonstrates good compression capacity of complex depth images to a small latent code, it still faces limitations in producing reconstructions that can be used to derive an accurate collision representation especially in cluttered and complex scenes. As expected, aggressive compression leads to loss of important information. However, compared to the depth image, a collision image would typically contain less complex and more low-frequency information regarding the same scene owing to the ``inflation'' of the obstacles. Due to this process, pixels corresponding to thin features in a depth image end up being represented by a larger region of pixels showing collision-free distance values. It is noted, transforming the depth image to a collision image requires a spatial understanding of the scene as robot size-inflated regions in the collision image occlude the regions near the edges of obstacles represented in depth images. Nonetheless, once a network is trained to predict this, it also implies reduction in the information that has to be kept during compression.

\begin{figure}[h]
    \centering
    \includegraphics[width=0.95\columnwidth]{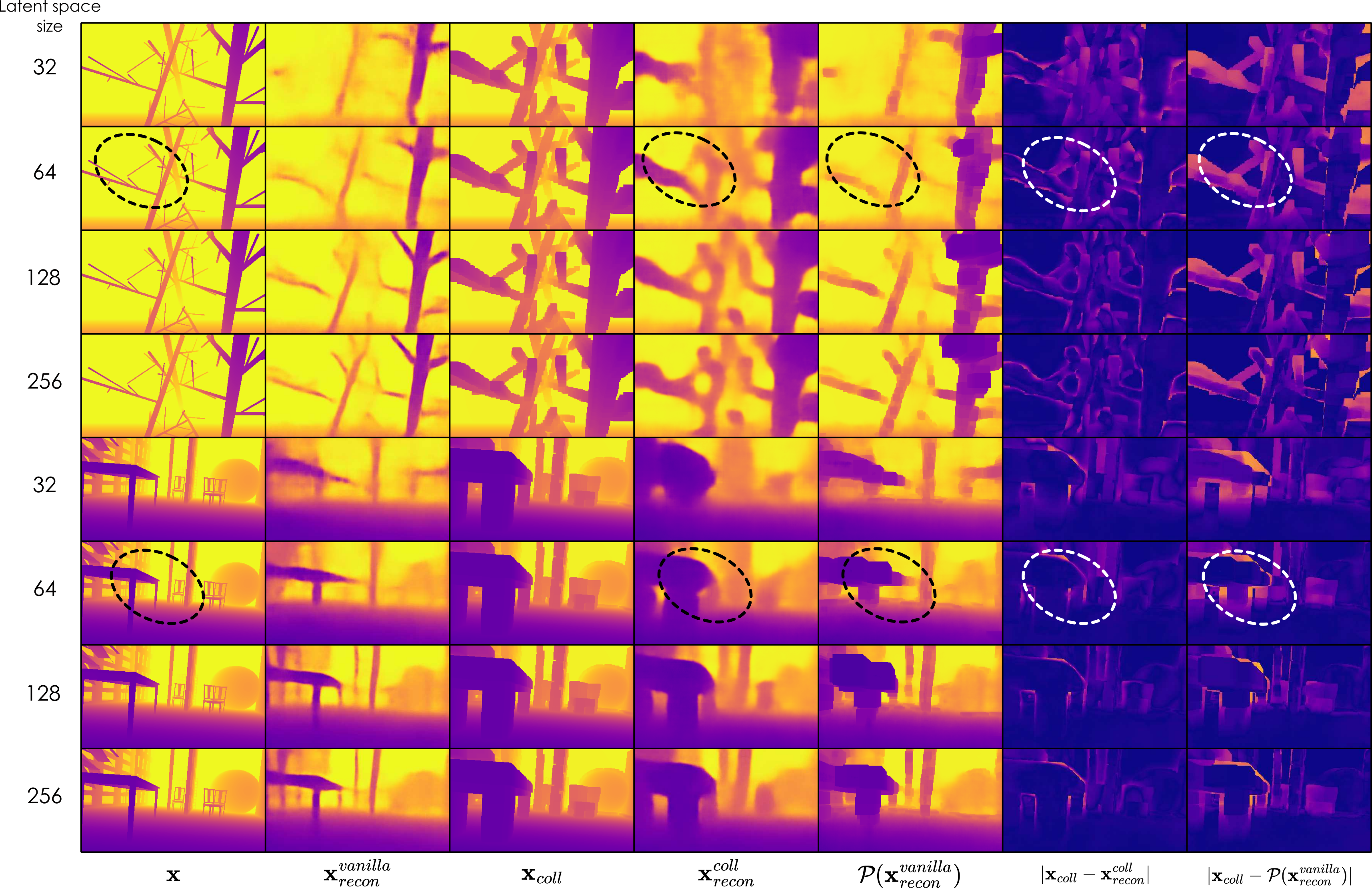}
    \vspace{-2ex}
    \caption{Comparison between \ac{collnet} and vanilla-VAE to derive the collision information from the input depth image for different levels of compression. The images compressed and reconstructed using \ac{collnet} ($\x^{coll}_{recon}$), vanilla-VAE ($\x^{vanilla}_{recon}$) are shown. A collision image $\mathcal{P}(\x^{vanilla}_{recon})$ is derived from $\x^{vanilla}_{recon}$. The derived collision images are compared against the ground-truth collision image $\xcoll$ for errors.}
    \label{fig:CollisionVAE_vs_vanilla_VAE}
    \vspace{-3ex}
\end{figure}

We compare the performance of the proposed \ac{collnet} against the task agnostic vanilla-VAE to compare the capability of these networks in retaining collision prediction information in the compressed latent space spanning from the depth image. The \ac{collnet} is trained to directly reconstruct the collision image, while the vanilla-VAE is trained to reconstruct the input depth image and thus for the purposes of assessing its capacity to retain the information needed for collision prediction, a new collision image is derived (as in Section~\ref{sec:proposed_method}) from the images reconstructed from its latent space through the decoder. Essentially, to ensure a fair comparison, we use the mapping $\mathcal{P}(\textbf{x}^{vanilla}_{recon})$ to obtain the derived collision image from the reconstructed input depth image. Figure~\ref{fig:CollisionVAE_vs_vanilla_VAE} presents examples of images reconstructed using both the \ac{collnet} and vanilla-VAE. 

The reconstructed collision image $\x^{coll}_{recon}$ and the derived collision image from the vanilla-VAE reconstruction $\mathcal{P}(\x^{vanilla}_{recon})$ are compared against the true collision image. The areas of errors are highlighted in Figure~\ref{fig:CollisionVAE_vs_vanilla_VAE}. The collision image derived from the vanilla-VAE reconstruction shows a greater number of regions with erroneous collision information, while the image from the \ac{collnet} $\x^{coll}_{recon}$ shows both smaller error magnitudes and only small regions of error when compared to the true collision image $\xcoll$. Moreover, the reconstructed collision image captures thin features such as branches in the environment and reconstruct the regions of collisions in the same. The results calculating the MSE of the reconstructed collision image and the derived collision image from the depth image reconstruction are presented in Table~\ref{tab:comparison_task_driven_compression}. As presented, the task-driven \ac{collnet} outperforms the vanilla-VAE by a large margin.

\begin{table}
    \vspace{-2ex}
    \centering
    \label{tab:comparison_task_driven_compression}
    \caption{Comparison of MSE for reconstructed images with \ac{collnet} and a transformed collision representation of the image reconstructed using vanilla-VAE.}
    \vspace{-2ex}
    \begin{tabularx}{\columnwidth}{|l|X|X|X|X|}
      \hline
      \multicolumn{5}{|c|}{\textbf{MSE against Collision Image $\xcoll$}} \\
      \hline
      \textbf{Latent dims:} & \textbf{32} & \textbf{64} & \textbf{128} & \textbf{256}\\
      \hline
      $\x^{coll}_{recon}$ & $\textbf{783.718}$ & $\textbf{516.487}$ & $\textbf{418.03}$ & $\textbf{402.66}$ \\
      \hline
      $\mathcal{P}(\x^{vanilla}_{recon})$ & $4828.50$ & $4339.76$ & $2532.14$ & $2539.89$ \\
      \hline
    \end{tabularx}
    \vspace{-3ex}
\end{table}

\section{Conclusions and Future Work} \label{sec:conclusions}
\vspace{-1ex}
This paper presented a learning-based method for task-driven aggressive compression of depth images to a highly compressed latent representation tailored to infer collision-free travel distances for a robot in the environment. A novel method was proposed to generate robot size-specific collision prediction data from given depth images using rendering frameworks. Such depth and collision prediction image tuples are then used to train a neural network performing the task-driven compression of encoding a latent space that captures collision information from depth images. We show that our proposed approach is able to encode depth images by a compression factor over $4000:1$, while retaining the information necessary to predict collisions from depth images of complex cluttered scenes. Moreover, we show that such purposeful neural network-based compression techniques demonstrate superior performance against traditional methods using FFT and wavelets or even conventional variational autoencoders for image reconstruction from highly compressed latent spaces.

\bibliographystyle{splncs04}
\bibliography{references}

\end{document}